\pdfoutput=1
\documentclass[11pt]{article}

\usepackage[preprint]{acl}
\usepackage{times}
\usepackage{epsfig}
\usepackage{graphicx}
\usepackage{amsmath}
\usepackage{amssymb}
\usepackage{url}
\usepackage{booktabs}
\usepackage{comment}
\usepackage{threeparttable}

\usepackage{multirow}
\usepackage{subcaption}
\usepackage[export]{adjustbox}
\usepackage[utf8]{inputenc} 
\usepackage[T1]{fontenc}    
\usepackage{amsfonts}       
\usepackage{nicefrac}       
\usepackage[misc]{ifsym}
\usepackage{mathtools}
\usepackage{amsthm}
\usepackage{enumitem}
\usepackage{pifont} 
\usepackage{longtable} 
\usepackage{bbm}

\title{T3D: Advancing 3D Medical Vision-Language Pre-training by Learning Multi-View Visual Consistency}

\author{
Che Liu\textsuperscript{1}, 
Cheng Ouyang\textsuperscript{2}, 
Yinda Chen\textsuperscript{5}, 
César Quilodrán-Casas\textsuperscript{1},
Lei Ma\textsuperscript{4}, 
Jie Fu\textsuperscript{3}, \\
\bf Yike Guo\textsuperscript{3}, 
\bf Anand Shah\textsuperscript{1}, 
\bf Wenjia Bai\textsuperscript{1}, and 
\bf Rossella Arcucci\textsuperscript{1} \\
\textsuperscript{1}Imperial College London, UK,
\textsuperscript{2}University of Oxford, UK, \\
\textsuperscript{3}Hong Kong University of Science and Technology, China,
\textsuperscript{4}Peking University, China, \\
\textsuperscript{5}University of Science and Technology of China, China \\
\href{mailto:che.liu21@imperial.ac.uk}{che.liu21@imperial.ac.uk} \\
}
\begin{document}
\maketitle

\begin{abstract}
While 3D visual self-supervised learning (vSSL) shows promising results in capturing visual representations, it overlooks the clinical knowledge from radiology reports. Meanwhile, 3D medical vision-language pre-training (MedVLP) remains underexplored due to the lack of a large-scale, publicly available 3D medical image-report dataset. To bridge this gap, we introduce \textbf{CT-3DVLP}, the first and largest \textbf{public} 3D volume-report dataset, establishing a comprehensive benchmark for 3D MedVLP research. Meanwhile, we propose the \textbf{T3D} framework, which enhances 3D MedVLP beyond naive CLIP-style alignment that directly pairs volumes with reports but neglects local visual representations. Instead, we introduce \textbf{Text-informed Multi-view Alignment (TMA)}, a novel approach that clusters volumetric data while enforcing consistency across different views of the same volume-report pair. TMA integrates textual features into fine-grained visual representations, ensuring contextual coherence across views.  We evaluate T3D across multiple downstream tasks in both unimodal and cross-modal settings, including zero-shot and fine-tuned classification, cross-modal retrieval, report generation, and semantic segmentation. Our results show that T3D consistently outperforms existing vSSL and multimodal methods, demonstrating superior zero-shot and fine-tuning capabilities and setting a new benchmark for 3D medical image understanding\footnote{All data and code will be released upon acceptance.}.
\end{abstract}

\section{Introduction}
Deep learning (DL) has transformed 3D medical image analysis, improving diagnostic accuracy and efficiency. However, supervised DL methods require extensive, high-quality annotations, which are costly and time-consuming. To reduce this dependency, visual self-supervised learning (vSSL) has shown great potential in leveraging large-scale unlabeled medical data. Existing vSSL techniques, including image restoration (IR) and contrastive learning (CL) \cite{chaitanya2020contrastive,taleb20203d,xie2022unimiss,haghighi2022dira}, have demonstrated effectiveness in learning visual representations.  

\begin{figure}
    \centering
    \includegraphics[width=0.99\linewidth]{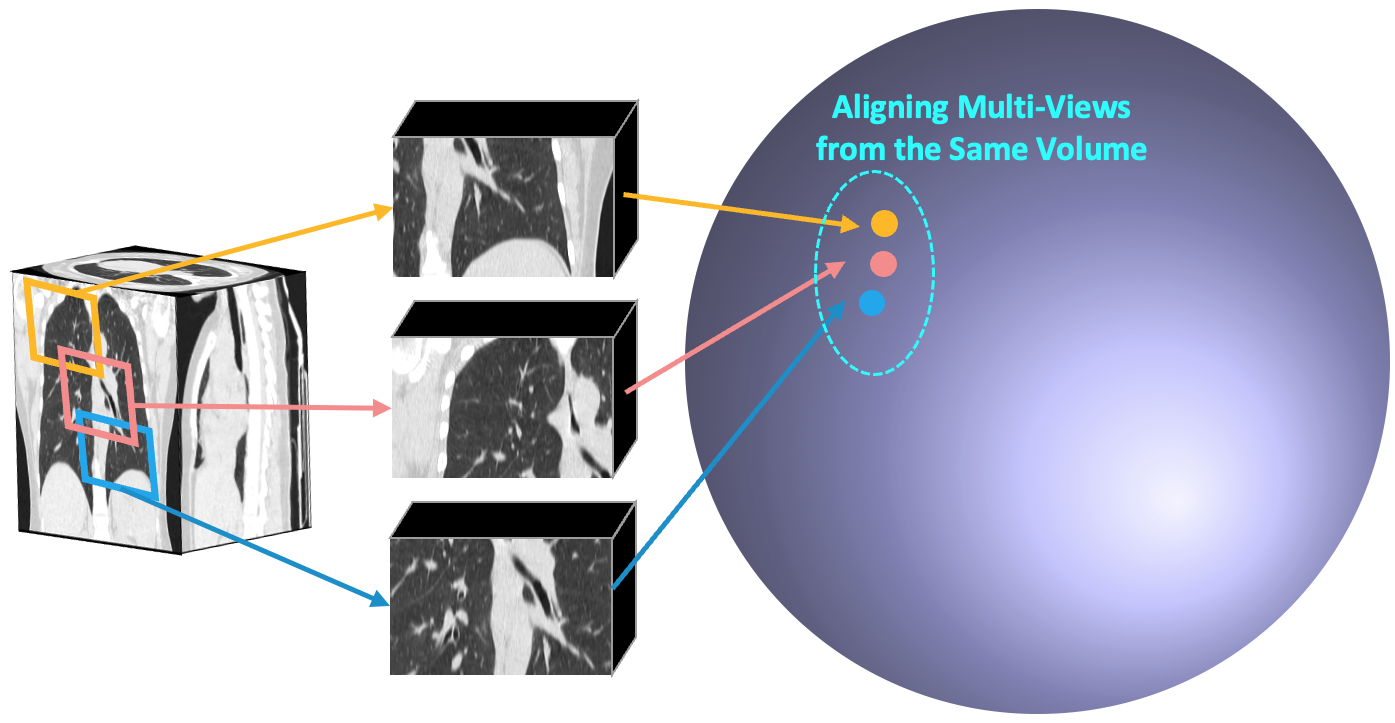}
    \caption{
      Illustration of the Text-Informed Multi-View Alignment (TMA) method. Multiple local views \(V_i^m\) are generated from the same 3D volume, and their embeddings are aligned in the latent space to encourage consistency across views from the same volume-report pair. Each view's embedding is refined by the corresponding report to ensure consistency among all views from the same volume. The details are illustrated in Section~\ref{sec:tma}.
    }
    \label{fig:tma}
\end{figure}

IR-based methods reconstruct images from corrupted versions \cite{vincent2010stacked,pathak2016context,chen2020generative,he2022masked,wei2022masked,xie2022simmim,gidaris2018unsupervised} but primarily capture low-level features, often overlooking high-level semantics crucial for tasks like disease classification and tumor segmentation \cite{liu2023improving,liu2023pixmim,he2022masked}. Meanwhile, CL-based vSSL methods \cite{swinunter,pcrlv2,vox2vec} enforce feature similarity between patches from the same image while treating patches from different images as negatives. However, these approaches risk semantic misalignment, as positive pairs may originate from anatomically distinct regions, while negative pairs might share similar structures \cite{alice}, leading to suboptimal feature learning.  

Medical Vision-Language Pre-training (MedVLP) has emerged as a promising approach to enhance representation learning by aligning medical images with radiology reports, providing clinically relevant supervision and improving feature informativeness in 2D medical imaging tasks \cite{radford2021learning,liu2023imitate,mgca,medunic}. However, its application to 3D medical images remains underexplored due to dataset scarcity and the lack of large-scale public benchmarks. Moreover, naive CLIP-style alignment relies solely on language supervision at the whole-volume level, limiting its ability to capture fine-grained 3D visual features.

To address these challenges, we propose \textbf{T3D}, a framework designed to enhance 3D MedVLP. Our key contributions include:  
\begin{itemize}  
    \item We propose \textbf{T3D}, which integrates Global Cross-modal Alignment (GCA) and \textbf{Text-informed Multi-view Alignment (TMA)}, a novel mechanism that refines visual representations by leveraging text-informed guidance to enforce consistency across different views while capturing fine-grained visual features.  

    \item To train T3D, we curate \textbf{CT-3DVLP}, the first large-scale public dataset including 52,639 paired CT volumes and radiology reports, establishing a comprehensive benchmark for 3D MedVLP research.

    \item Benefiting from the novel alignment, T3D demonstrates superior performance across various downstream tasks in both unimodal and cross-modal settings, including zero-shot and fine-tuned classification, retrieval, report generation, and segmentation.
 
\end{itemize}

\section{Related Work}

\noindent\textbf{VLP for 2D Medical Images\hspace{1mm}}  
VLP has been extensively explored for 2D medical imaging to bridge visual and textual modalities. Early works such as ConVIRT~\cite{convirt} introduced global image-text alignment, later refined by GLoRIA and MGCA~\cite{huang2021gloria,mgca}, which incorporated local alignment for better cross-modal representation learning. Other methods, including Med-UniC~\cite{medunic}, mitigated language biases, while MedKLIP~\cite{medklip} and KAD~\cite{kad} leveraged domain-specific knowledge. Additionally, reconstruction-based approaches like MRM~\cite{zhouadvancing} and PRIOR~\cite{prior} utilized image-text token prediction tasks, further improved by \cite{huang2023enhancing} through adaptive token weighting.  
Despite these advancements, 2D VLP methods do not directly transfer to 3D imaging. The volumetric nature of 3D data introduces challenges in aligning 3D scans with textual reports due to high computational costs.
While patch-based methods~\cite{swinunter,vox2vec,alice} attempt to retain local information, they often lead to misalignment between cropped sub-volumes and full medical reports. These limitations highlight the need for specialized VLP approaches tailored for 3D medical imaging.

\noindent\textbf{VLP for 3D Medical Images\hspace{1mm}}  
While VLP has advanced general 3D vision~\cite{ulip,zeng2023clip2,ulip2,chen2023end}, these methods focus on sparse 3D point clouds and are not directly applicable to dense medical volumes like CT scans. Early 3D MedVLP approaches~\cite{gtgm,medblip} attempted to align full medical reports with cropped sub-volumes, introducing misalignment biases. To address this issue, \cite{wu2023towards,lei2023unibrain} proposed downsampling high-resolution volumes for report alignment, but this leads to a loss of anatomical details crucial for segmentation and diagnosis.  

Additionally, most 3D MedVLP works rely on private datasets~\cite{cao2024bootstrapping,shui2025large}, limiting reproducibility. They also heavily depend on external annotation tools, such as segmenting each anatomical region and categorically labeling volumes, which introduces additional annotation costs and potential inconsistencies. These limitations underscore the need for a publicly available dataset to advance open 3D MedVLP research.

\noindent\textbf{vSSL for 3D Medical Imaging\hspace{1mm}}  
vSSL has been widely explored in 3D medical imaging, with image restoration (IR) and contrastive learning (CL) as dominant strategies. IR-based methods reconstruct corrupted images~\cite{he2022masked,wei2022masked,xie2022simmim} but primarily capture low-level features, often overlooking high-level semantics crucial for diagnosis~\cite{liu2023improving,liu2023pixmim,he2022masked}. While recent works~\cite{patchswap,mg,wu2024voco} incorporated anatomical priors, comprehensive semantic understanding remains underexplored. CL-based methods~\cite{chaitanya2020contrastive,taleb20203d,xie2022unimiss} enforce similarity between patches from the same image while treating patches from different images as negatives. However, positive pairs may originate from distinct anatomical regions, leading to semantic misalignment~\cite{alice}. vox2vec~\cite{vox2vec} introduced voxel-level alignment, but treating neighboring voxels as negatives introduces bias.  
Despite advancements, existing vSSL methods lack clinical knowledge integration, limiting their effectiveness in real-world medical tasks.

\section{Method}
\subsection{Extracting Visual and Text Features}
\begin{figure*}[ht!]
    \centering
    \includegraphics[width=0.99\linewidth]{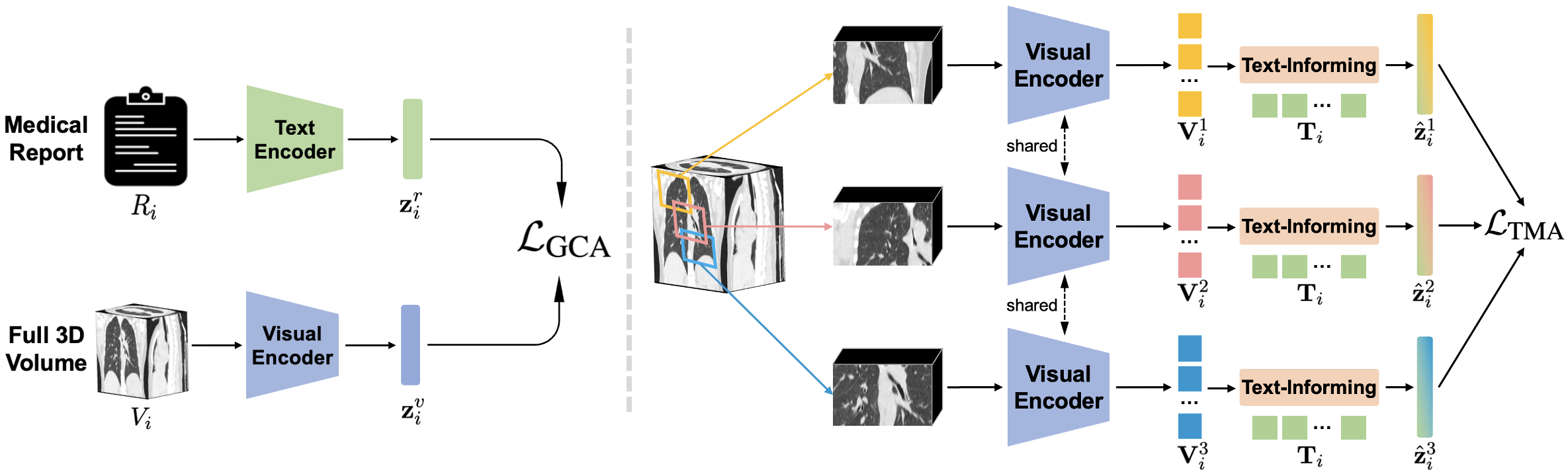}
    \caption{
        The T3D framework for learning multi-level 3D visual representations from corresponding medical reports.  
        \textbf{Left:} To learn global cross-modal representations, we align the full 3D volume \(V_i\) with its corresponding medical report \(R_i\) using the loss function \(\mathcal{L}_{\textrm{GCA}}\). The output embeddings \(\mathbf{z}_i^v\) and \(\mathbf{z}_i^r\) are optimized to encourage the matching of paired visual and textual features.  
        \textbf{Right:} To further capture fine-grained visual representations, we first generate \(M\) local views \(V_i^m\) from the same volume using random cropping. The same visual encoder, as used in the GCA framework, is applied to obtain the embeddings for these local views. We then refine these embeddings using the report embedding \(\mathbf{T}_i\), encouraging the local views from the same volume-report pair to become more similar in the latent space by minimizing the loss \(\mathcal{L}_{\textrm{TMA}}\).
        }
    \label{fig:frame}
\end{figure*}

Let $\mathcal{D} = \{(V_i, R_i)\}_{i=1}^{N}$ be a dataset of $N$ samples, where each 3D medical image 
\[
V_i \in \mathbb{R}^{1 \times H \times W \times S}
\]
(with height $H$, width $W$, and slices $S$) is paired with a radiology report $R_i$. 
We define a 3D visual encoder \( f_\theta(\cdot) \) (e.g., 3D ResNet-50) that maps an input volume \( V_i \) to a latent feature map \( F_i \):

\[
F_i = f_\theta(V_i) \in \mathbb{R}^{d_f \times h \times w \times s}
\]

where \( d_f \) is the number of output channels, and \( h, w, s \) are spatial dimensions. A global 1D embedding is obtained via average pooling over all spatial dimensions, followed by a learnable linear projection:

\[
\mathbf{z}^v_i = P^v\bigl(\mathrm{AvgPool}(F_i)\bigr) \in \mathbb{R}^{768}.
\]

For text, we use Med-CPT \cite{jin2023medcpt} as the text encoder \( g_\phi(\cdot) \) to obtain token embeddings from the radiology report \( R_i \):

\[
\mathbf{T}_i = g_\phi(R_i) \in \mathbb{R}^{L_r \times d_r},
\]

where \( L_r \) is the token length and \( d_r \) is the embedding dimension. We extract the [CLS] token embedding and project it into the shared space:

\[
\mathbf{z}^r_i = P^r\bigl(\mathbf{t}_i^{[\mathrm{CLS}]}\bigr) \in \mathbb{R}^{768}, \quad \text{where} \quad \mathbf{t}_i^{[\mathrm{CLS}]} \in \mathbb{R}^{d_r}.
\]

\subsection{Global Cross-Modal Alignment}
To learn the global cross-modal representation, we align 3D volumes with their corresponding reports using a CLIP loss, as shown in the left part of Figure \ref{fig:frame}. Given a batch of \( B \) samples, the loss \( \mathcal{L}_{\textrm{GCA}} \) is defined as:
\[
\mathcal{L}_{\textrm{GCA}} =
 -\sum_{i=1}^B \log 
    \frac{
      \exp\bigl(\mathrm{sim}(\mathbf{z}^v_i, \mathbf{z}^r_i) / \tau\bigr)
    }{
      \sum_{j=1}^B \exp\bigl(\mathrm{sim}(\mathbf{z}^v_i, \mathbf{z}^r_j) / \tau\bigr)
    },
\]

where \( \mathrm{sim}(\cdot,\cdot) \) denotes the dot product similarity, and we set \( \tau = 0.07 \) following~\cite{radford2021learning}.

\subsection{Text-Informed Multi-View Alignment}
\label{sec:tma}
\paragraph{Motivation.}
While the CLIP loss aligns 3D volumes with their corresponding radiology reports at a global level, it fails to capture fine-grained visual features crucial for understanding 3D medical imaging. To address this, we propose a text-informed multi-view alignment scheme that encourages consistency across multiple local 3D sub-volumes paired with the same report, as illustrated in Figure \ref{fig:tma}.

\paragraph{Generating Local Views.}
To enable the learning of fine-grained visual features, we generate multiple 3D local views from each \(V_i\) by randomly cropping\footnote{We implement the random cropping using the MONAI package \url{https://docs.monai.io/en/stable/transforms.html}} sub-volumes of size \(128 \times 128 \times 64\):
\[
\{V_i^m\}_{m=1}^M = \mathrm{RandomCrop}\bigl(V_i, 128 \times 128 \times 64\bigr),
\]
\[
V_i^m \in \mathbb{R}^{1 \times 128 \times 128 \times 64}.
\]
Passing each \(V_i^m\) through the 3D encoder \(f_\theta(\cdot)\) yields the corresponding feature map:

\[
F_i^m = f_\theta\bigl(V_i^m\bigr) \in \mathbb{R}^{d_f \times h' \times w' \times s'},
\]
\[
h' < h, \quad w' < w, \quad s' < s.
\]

\paragraph{Text-Informed Local Feature Enhancement.}  
To mitigate biases in local view alignment, we incorporate text-informed features into each local 3D view. Treating views from different volumes as negatives can be problematic, as they may contain similar anatomical regions. Similarly, views from the same volume may originate from distinct regions, making naïve positive pairing unreliable. 

Since each volume is paired with a unique radiology report, we leverage textual information to refine local visual representations, as shown in the right part of Figure \ref{fig:frame}. This integration ensures that semantically similar regions across different volumes are not misclassified as negatives while refining positive associations within the same volume. By conditioning local views on text, we reduce bias in positive and negative pair selection before alignment.

We extract text token embeddings \(\mathbf{T}_i\) from the text encoder \(g_\phi(\cdot)\) and reshape the local 3D feature map \(F_i^m\) into a sequence:
\[
\mathbf{V}_i^m = \text{Reshape}(F_i^m) \in \mathbb{R}^{L_v \times d_f},
\]
\[
L_v = h' \times w' \times s'.
\]
A single-layer Transformer block \(\mathcal{F}_\psi(\cdot)\) refines \(\mathbf{V}_i^m\) using \(\mathbf{T}_i\) as keys and values, followed by average pooling to obtain the text-informed local embedding:
\[
\hat{\mathbf{z}}_i^m = \mathrm{AvgPool} \bigl( \mathcal{F}_\psi (\mathbf{V}_i^m, \mathbf{T}_i) \bigr) \in \mathbb{R}^{d_f}.
\]

\paragraph{Multi-View Alignment.}  
Since multiple local views \(\{V_i^m\}_{m=1}^M\) are generated per volume-report pair, a naïve contrastive loss is unsuitable as it assumes one-to-one positive pairings. Instead, we assign each local view to one of \( B \) cluster labels, where \( B \) is the batch size, and each cluster label corresponds to a specific volume-report pair in the batch. A linear projection layer is used to predict the cluster assignment probability:

\[
\mathcal{L}_{\textrm{TMA}} = -\sum_{i=1}^{B} \sum_{m=1}^{M} \log 
\frac{\exp\bigl(f(\hat{\mathbf{z}}_i^m)_{c_i} / \tau\bigr)}
{\sum_{j'} \exp\bigl(f(\hat{\mathbf{z}}_i^m)_{j'} / \tau\bigr)}.
\]

where \( f(\cdot) \) is a linear projection function, and \( c_i \) is the assigned cluster label for the corresponding volume-report pair. This objective encourages views from the same pair to cluster together while distinguishing them from those of different pairs.

\subsection{Overall Objective}
The final optimization objective of T3D aims to learn both global and local representations through:
\[
\mathcal{L}_{\textrm{total}} = \mathcal{L}_{\textrm{GCA}} + \mathcal{L}_{\textrm{TMA}}.
\]

\section{Experiments}
\paragraph{Pre-training Dataset.}
To construct the largest publicly available dataset for 3D MedVLP, we curate data from three public resources: BIMCV-R~\cite{chen2024bimcv}, CT-RATE~\cite{ct-rate}, and INSPECT~\cite{huang2023inspect}. We include all available samples from these datasets for pretraining, except for the official test set of CT-RATE. Additionally, we split the test set from BIMCV-R for later cross-modal tasks, following \cite{chen2024bimcv}. For preprocessing, we follow the RadGenome-CT~\cite{zhang2024radgenome} pipeline to extract 
unique CT-report pairs. In total, we obtain 52,639 samples, with 6,548 samples from BIMCV-R, 25,691 samples from CT-RATE, and 20,400 samples from INSPECT for pre-training. All CT volumes are resampled to a spacing of [1,1,4] mm, resized to $256 \times 256 \times 128$, and normalized to the range [0,1] after truncating Hounsfield unit (HU) values to $[-1000, +1000]$.

\paragraph{Pre-training Implementation.}
We use a 3D ResNet50\footnote{We use the implementation from \url{https://docs.monai.io/en/stable/networks.html}} as the visual encoder and MedCPT~\cite{jin2023medcpt} as the text encoder. 
The AdamW optimizer is employed with a learning rate of $1 \times 10^{-3}$ and a cosine annealing scheduler. We pretrain for 50 epochs with a 5-epoch warmup. The batch size is set to 32 per GPU, and we implement our training on 8 A100-80G GPUs, resulting in a total effective batch size of 256. 
No data augmentation is applied to the volumes to preserve spatial integrity and intensity. We only use random cropping to generate three local views of size $128 \times 128 \times 64$. In this study, we generate three local views for each 3D volume-report pair.

\subsection{Downstream Tasks Configuration}
We evaluate T3D on a variety of downstream tasks:

\noindent\textbf{Classification:} We assess zero-shot and fine-tuned classification on the CT-RATE~\cite{ct-rate} and CC-CCII~\cite{zhang2020clinically} official test set.

\noindent\textbf{Cross-modal Retrieval:} We evaluate zero-shot image-to-text and text-to-image retrieval on the BIMCV-R dataset following~\cite{chen2024bimcv}.

\noindent\textbf{Report Generation:} We evaluate this task on the official test set of CT-RATE~\cite{ct-rate}, implementing it based on the LLaVA architecture~\cite{liu2024visual}. We use Qwen2.5-7B-Instruct \footnote{https://huggingface.co/Qwen/Qwen2.5-7B-Instruct} as the LLM backbone and employ the visual encoder from our work and baselines to extract the visual representation.

\noindent\textbf{Segmentation:} We perform multi-organ segmentation on the AMOS~\cite{ji2022amos} dataset and lung tumor segmentation on the MSD-Lung~\cite{antonelli2022medical} dataset, following the protocols in VoCo~\cite{wu2024large,wu2024voco}.

The detailed configurations and implementations for these downstream tasks are provided in Appendix~\ref{sec:downstream detail}.

\subsection{Baseline Selection}
We compare T3D with several state-of-the-art (SOTA) visual representation learning methods via vSSL and language supervision:

\noindent\textbf{3DMAE~\cite{mae3d}:} A vSSL-based model that reconstructs pixel-level features from masked volumes to learn low-level visual representations.

\noindent\textbf{VoCo~\cite{wu2024voco}:} A vSSL-based model that crops sub-volumes and predicts their locations in the original volume, learning relative local visual features.

\noindent\textbf{MRM~\cite{zhouadvancing}:} A 2D MedVLP method that applies masked image and text modeling, leveraging cross-modal reconstruction to learn joint representations. 

\noindent\textbf{IMITATE~\cite{liu2023imitate}:} A 2D MedVLP method that aligns multi-level visual features with different sections of the report.

\noindent\textbf{CT-CLIP~\cite{ct-rate}:} A 3D MedVLP model that aligns the entire volume with text using the original CLIP loss.

\noindent\textbf{Merlin~\cite{blankemeier2024merlin}:} A 3D MedVLP model trained on in-house data using CLIP-style alignment.

For 3DMAE \cite{mae3d}, we use their official code to reimplement them on our curated dataset since they do not release official pretrained weights. For MRM~\cite{zhouadvancing} and IMITATE~\cite{liu2023imitate}, we replace their 2D visual encoder with a 3D version for a fair comparison and use 3D input. 
For VoCo \cite{wu2024voco}, Merlin \cite{blankemeier2024merlin}, and CT-CLIP \cite{ct-rate}, we use their official pretrained weights to ensure a fair comparison.

\section{Results}
We evaluate T3D across a range of downstream tasks, comparing its performance against several state-of-the-art models. Our results are presented for each task in terms of standard evaluation metrics, and we discuss the performance of T3D in comparison to the baselines.
\begin{table*}[ht]
    \centering
    \resizebox{0.99\textwidth}{!}{
    \begin{tabular}{l|cccc|cccc|cccc|cccc|ccc|ccc}
        \toprule
        \multirow{3}{*}{\textbf{Method}} 
        & \multicolumn{8}{c|}{\textbf{Zero-Shot Classification}} 
        & \multicolumn{8}{c|}{\textbf{Fine-Tune Classification}} 
        & \multicolumn{6}{c}{\textbf{Cross-Modal Retrieval}} \\
        \cmidrule(lr){2-9} \cmidrule(lr){10-17} \cmidrule(lr){18-23}
        & \multicolumn{4}{c|}{\textbf{CT-RATE}} 
        & \multicolumn{4}{c|}{\textbf{CC-CCII}} 
        & \multicolumn{4}{c|}{\textbf{CT-RATE}} 
        & \multicolumn{4}{c|}{\textbf{CC-CCII}}
        & \multicolumn{3}{c|}{\textbf{BIMCV-R (Text to Image)}} 
        & \multicolumn{3}{c}{\textbf{BIMCV-R (Image to Text)}} \\
        \cmidrule(lr){2-5} \cmidrule(lr){6-9}
        \cmidrule(lr){10-13} \cmidrule(lr){14-17}
        \cmidrule(lr){18-20} \cmidrule(lr){21-23}
        & Prec. & AUC & ACC & F1
        & Prec. & AUC & ACC & F1
        & Prec. & AUC & ACC & F1
        & Prec. & AUC & ACC & F1
        & R@1 & R@5 & R@10 & R@1 & R@5 & R@10 \\
        \midrule
        \multicolumn{23}{c}{\textbf{Visual SSL only}} \\
        \midrule
        3DMAE \cite{mae3d}
            & / & / & / & /
            & / & / & / & /
            & 30.1 & 70.4 & 64.7 & 64.8
            & 82.7 & 88.4 & 87.6 & 85.4
            & / & / & / & / & / & / \\
        VoCo \cite{wu2024voco}
            & / & / & / & /
            & / & / & / & /
            & 32.0 & 72.0 & 68.1 & 69.4
            & 87.9 & 90.9 & 90.83\(^*\) & 88.7
            & / & / & / & / & / & / \\
        \midrule
        \multicolumn{23}{c}{\textbf{Language Supervision}} \\
        \midrule
        MRM \cite{zhouadvancing}
            & 27.6 & 67.3 & 61.4 & 65.2 
            & 65.2 & 82.1 & 78.5 & 80.0
            & 32.4 & 74.8 & 67.5 & 68.6
            & 85.2 & 90.7 & 88.0 & 88.5
            & 3.0 & 7.2 & 21.3 & 3.2 & 7.6 & 20.9 \\
        IMITATE \cite{liu2023imitate}
            & 29.5 & 68.9 & 63.6 & 66.4 
            & 68.6 & 83.7 & 80.2 & 81.5
            & 33.0 & 74.3 & 68.2 & 69.7
            & 86.4 & 91.5 & 89.2 & 89.7
            & 3.1 & 7.9 & 21.5 & 3.6 & 7.8 & 21.7 \\
        CT-CLIP \cite{ct-rate}
            & 30.6\(^*\) & 70.4\(^*\) & 65.1\(^*\) & 69.1\(^*\)
            & 71.6 & 84.3 & 82.3 & 83.0
            & 34.2\(^*\) & 75.0\(^*\) & 69.2\(^*\) & 72.8\(^*\)
            & 90.8 & 92.0 & 91.4 & 90.3
            & 3.9 & 8.3 & 22.4 & 3.7 & 8.5 & 22.9 \\
        Merlin \cite{blankemeier2024merlin}
            & 33.7\(^*\) & 72.8\(^*\) & 67.2\(^*\) & 70.9\(^*\)
            & 73.2 & 86.4 & 85.0 & 85.9
            & 37.1 & 76.2 & 71.0 & 75.0
            & 91.5 & 91.9 & 91.5 & 89.6
            & 4.0 & 8.7 & 23.5 & 4.1 & 8.9 & 23.4 \\
        \midrule
        \textbf{T3D (Ours)}
            & \textbf{35.1} & \textbf{73.7} & \textbf{69.0} & \textbf{72.5}
            & \textbf{75.0} & \textbf{89.4} & \textbf{88.3} & \textbf{87.2}
            & \textbf{39.5} & \textbf{80.2} & \textbf{76.3} & \textbf{77.8} 
            & \textbf{93.1} & \textbf{93.2} & \textbf{92.7} & \textbf{92.1}
            & \textbf{4.7} & \textbf{10.0} & \textbf{25.6} & \textbf{4.9} & \textbf{10.4} & \textbf{25.9} \\
        \bottomrule
    \end{tabular}
    }
    \caption{Performance comparison of visual SSL and language supervision methods on zero-shot classification, fine-tune classification, and cross-modal retrieval tasks. `/' indicates that visual SSL methods are unable to perform cross-modal tasks since they only learn representations from images. `$^{*}$' denotes results directly cited from \cite{shui2025large,wu2024voco}. The best results in each column are highlighted in \textbf{bold}.}
    \label{tab:cls ret}
\end{table*}

\begin{table}[ht]
    \centering
    \resizebox{0.9\linewidth}{!}{
    \begin{tabular}{l|c|c}
        \toprule
        \multirow{2}{*}{\textbf{Method}} & \textbf{AMOS} & \textbf{MSD-Lung} \\
        \cmidrule(lr){2-3}
         &Dice  &Dice  \\
        \midrule
        \multicolumn{3}{c}{\textbf{Visual SSL only}} \\
        \midrule
        3DMAE \cite{mae3d} & 82.71$^{*}$ & 65.32 \\
        VoCo \cite{wu2024voco} &  88.06$^{*}$ & 68.99$^{*}$ \\
        \midrule
        \multicolumn{3}{c}{\textbf{Language Supervision}} \\
        \midrule
        MRM \cite{zhouadvancing} & 85.12 & 65.67 \\
        IMITATE \cite{liu2023imitate} & 84.51 & 67.31 \\
        CT-CLIP \cite{ct-rate}  & 83.44 & 68.37 \\
        Merlin \cite{blankemeier2024merlin} & 84.74 & 68.89 \\
        \midrule
        \textbf{T3D (Ours)} & \textbf{89.83} & \textbf{70.12} \\
        \bottomrule
    \end{tabular}
    }
    \caption{Semantic segmentation performance comparison of visual SSL and language supervision methods on AMOS and MSD-Lung datasets. Dice scores are reported for both datasets. `$^{*}$' denotes results directly cited from \cite{wu2024voco}. The best results in each column are highlighted in \textbf{bold}.}
    \label{tab:seg_results}
\end{table}

\begin{table*}[ht]
    \centering
    \resizebox{0.99\linewidth}{!}{
    \begin{tabular}{l|ccccccc|ccccc}
        \toprule
        \multirow{3}{*}{\textbf{Method}} 
        & \multicolumn{12}{c}{\textbf{Report Generation on CT-RATE}} \\
        \cmidrule(lr){2-13}
        & \multicolumn{7}{c|}{\textbf{Lexical Metrics}} & \multicolumn{5}{c}{\textbf{Clinical Efficacy Metric}} \\
        \cmidrule(lr){2-8} \cmidrule(lr){9-13}
         & BLEU-1 & BLEU-2 & BLEU-3 & BLEU-4 & ROUGE-1 & ROUGE-2 & ROUGE-L
         & Precision & Recall & F1 & GREEN & RaTEScore \\
        \midrule
        \multicolumn{13}{c}{\textbf{Visual SSL only}} \\
        \midrule
        3DMAE \cite{mae3d}
        & 26.5 & 17.9 & 13.2 & 10.3 & 24.0 & 19.4 & 21.2
        & 13.1 & 9.4 & 11.0 & 26.9 & 35.3 \\
        VoCo \cite{wu2024voco}
        & 30.3 & 23.7 & 18.5 & 15.2 & 30.1 & 25.1 & 27.4
        & 16.3 & 13.2 & 14.6 & 31.7 & 39.6 \\
        \midrule
        \multicolumn{13}{c}{\textbf{Language Supervision}} \\
        \midrule
        MRM \cite{zhouadvancing}
        & 34.7 & 26.9 & 20.9 & 16.7 & 33.4 & 27.2 & 29.6
        & 20.4 & 15.3 & 17.5 & 36.7 & 44.9 \\
        IMITATE \cite{liu2023imitate}
        & 41.2 & 31.7 & 25.1 & 21.0 & 37.9 & 31.1 & 32.4
        & 25.8 & 17.2 & 20.7 & 40.2 & 49.1 \\
        CT-CLIP \cite{ct-rate}
        & 44.4 & 34.4 & 27.9 & 23.6 & 40.1 & 33.8 & 30.9
        & 31.7 & 18.1 & 25.3 & 42.9 & 53.1 \\
        Merlin \cite{blankemeier2024merlin}
        & 47.9 & 35.6 & 28.2 & 24.1 & 41.5 & 35.0 & 36.0
        & 33.1 & 19.3 & 25.8 & 46.4 & 56.5 \\
        \midrule
        \textbf{\textbf{T3D (Ours)}}
        & \textbf{50.1} & \textbf{38.3} & \textbf{30.4} & \textbf{26.2} & \textbf{43.8} & \textbf{36.7} & \textbf{37.8}
        & \textbf{35.5} & \textbf{20.7} & \textbf{27.4} & \textbf{49.2} & \textbf{59.6} \\
        \bottomrule
    \end{tabular}
    }
    \caption{Comparison of methods on the report generation task on the CT-RATE official test set using both lexical and clinical efficacy metrics. Lexical metrics include BLEU-1 to BLEU-4 and ROUGE-1, ROUGE-2, and ROUGE-L scores, while clinical metrics include Precision, Recall, and F1 following \cite{hamamci2024ct2rep}, as well as GREEN \cite{ostmeier2024green} and RaTEScore \cite{zhao2024ratescore}. The best results in each column are highlighted in \textbf{bold}.}
\label{tab:report_gen_results}
\end{table*}

\subsection{Zero-shot and Fine-tuned Classification}
For zero-shot classification and the fine-tuning setting, T3D outperforms all baselines, achieving the highest accuracy across both the CT-RATE \cite{ct-rate} and CC-CCII \cite{zhang2020clinically} datasets. Notably, it surpasses visual SSL methods and language supervision methods, demonstrating the superiority of our proposed framework in terms of precision, AUC, and F1-score, as shown in Table \ref{tab:cls ret}. Furthermore, in the fine-tuning setting, all language supervision baselines reach or even outperform the visual SSL methods \cite{mae3d,wu2024voco} across both datasets. This demonstrates the necessity of designing a VLP method to learn more representative 3D visual features.

\subsection{Cross-modal Retrieval}
In zero-shot cross-modal retrieval, we implement both image-to-text and text-to-image retrieval tasks to evaluate how well the image and text representations are aligned. T3D outperforms all baselines on the R@1, R@5, and R@10 metrics in both tasks. This demonstrates the superiority of our framework and the benefits of the multi-view alignment strategy in enhancing cross-modal representation learning.

\subsection{Report Generation}
For the report generation task, as shown in Table \ref{tab:report_gen_results}, we use lexical metrics such as BLEU-1 to BLEU-4 and ROUGE-1, ROUGE-2, and ROUGE-L to evaluate the quality of generated reports. Additionally, we utilize clinical efficacy metrics, including Precision, Recall, and F1, following \cite{hamamci2024ct2rep}, to assess the relevance and accuracy of the reports. On both types of metrics, T3D outperforms all baselines, demonstrating the superiority of our framework in generating high-quality reports across both lexical and clinical dimensions.

\begin{figure*}[ht]
    \centering
   \includegraphics[width=0.99\linewidth]{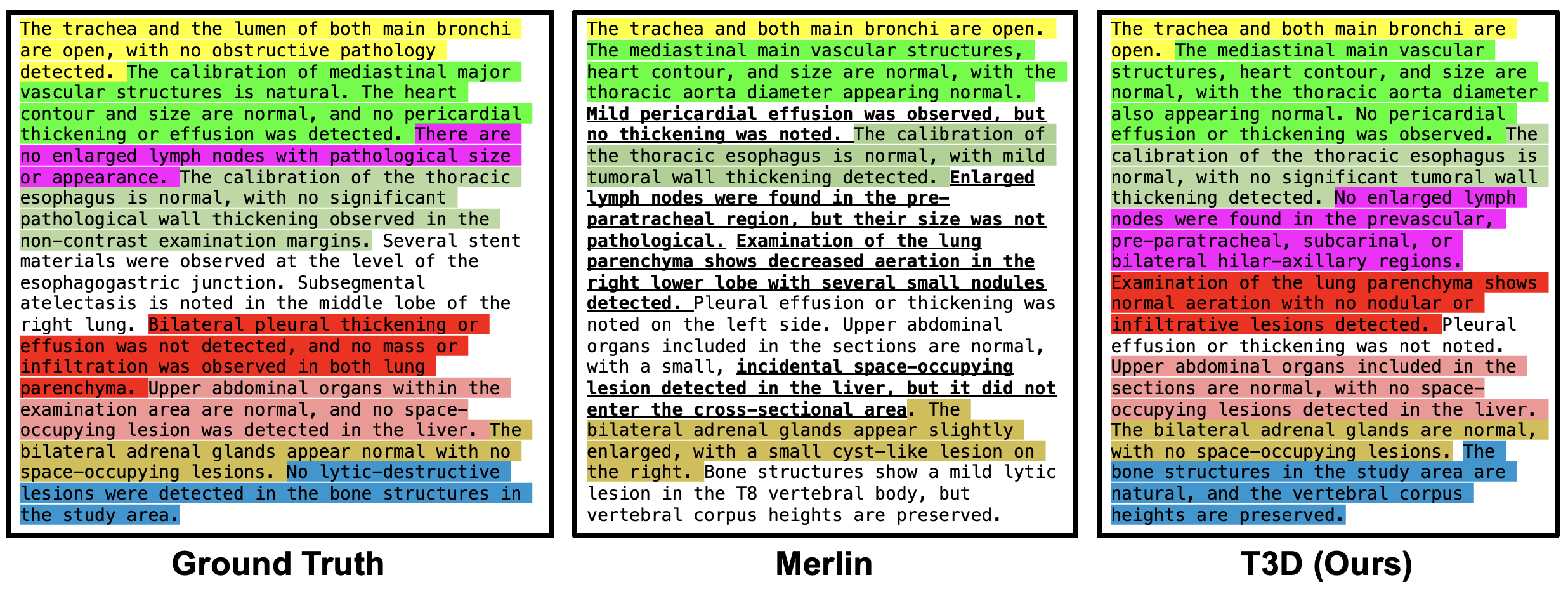}
    \caption{Report generation results of Merlin \cite{blankemeier2024merlin} and \textbf{T3D (Ours)}. Text highlighted in the same color indicates correct predictions, while bold and underlined text marks incorrect parts. Merlin shows incorrect patterns in various areas, whereas T3D provides more accurate results, particularly in the detection of lymph nodes and other pathologies.}
    \label{fig: report}
\end{figure*}

Furthermore, all language supervision-based visual encoders outperform the visual SSL-only methods, as shown in Table \ref{tab:report_gen_results}. This highlights that, on the report generation task, the multimodal representations learned through language supervision result in better performance, benefiting the task's specific requirements. Sample generated reports are shown in Figure \ref{fig: report}. As shown, our method detects the correct patterns, including subtle ones such as lymph nodes.

\subsection{Semantic Segmentation}
We further evaluate the dense visual representations learned from our method using multi-organ segmentation on the AMOS \cite{ji2022amos} and MSD-Lung tumor segmentation \cite{msd} datasets. The results, as shown in Table \ref{tab:seg_results}, demonstrate that our method, T3D, outperforms all other methods. Although VoCo \cite{wu2024voco} substantially outperforms other language supervision methods on the organ segmentation task, it does not achieve the same advantage on the tumor segmentation task. This highlights the limitations of visual SSL methods, which may not fully capture the complexities of tumor segmentation. However, our method, T3D, still surpasses VoCo \cite{wu2024voco}, which can be attributed to our multi-view alignment approach. This approach allows for learning multi-level visual features, significantly benefiting the segmentation task.

\begin{figure}[h]
    \centering
  \includegraphics[width=0.99\linewidth]{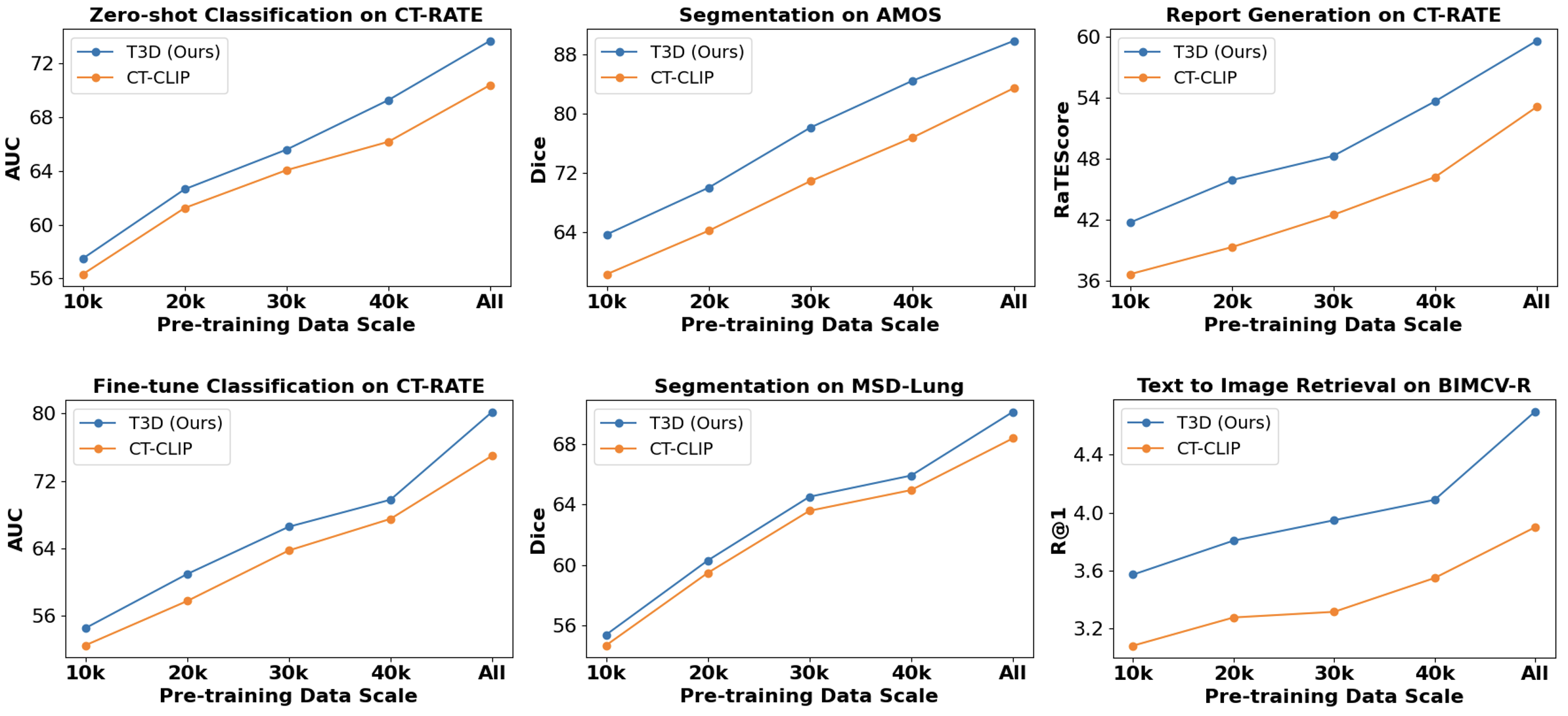}
    \caption{Comparison of T3D (Ours) and CT-CLIP \cite{ct-rate} across six tasks, showing AUC, Dice, RaTES score, and R@1 for varying pre-training data scales from 10k to the full dataset. T3D consistently outperforms CT-CLIP across all data scales and tasks, particularly with larger datasets.}
    \label{fig:scale}
\end{figure}

\section{Analysis}

\begin{table}[h]
    \centering
    \captionsetup{justification=centering}
    \subfloat[Loss Function Ablation.]{
        \resizebox{0.42\textwidth}{!}{
        \begin{tabular}{cc|ccc}
            \toprule
            \multicolumn{2}{c|}{\textbf{Loss Components}} & Zero-shot Classification & Segmentation & Report Generation \\
            \cmidrule(lr){1-2} \cmidrule(lr){3-5}
            $\mathcal{L}_{\textrm{GCA}}$ & $\mathcal{L}_{\textrm{TMA}}$  & CT-RATE (AUC) & AMOS (Dice) & CT-RATE (RaTEScore) \\
            \midrule
            \checkmark & & 71.8 & 85.0 & 56.3  \\
             & \checkmark & 71.4 & 85.2 & 55.9  \\
            \checkmark & \checkmark  & \textbf{73.7} & \textbf{89.8} & \textbf{59.6} \\
            \bottomrule
        \end{tabular}
        }
    }
\\
\vspace{5pt}
    \subfloat[Effect of Text-Informing on $\mathcal{L}_{\textrm{TMA}}$.]{
        \resizebox{0.42\textwidth}{!}{
        \begin{tabular}{c|ccc}
            \toprule
            \multirow{2}{*}{\textbf{$\mathbf{\mathcal{L}_{\textrm{TMA}}}$}} & Zero-shot Classification & Segmentation & Report Generation \\
            \cmidrule(lr){2-4}
             & CT-RATE (AUC) & AMOS (Dice) & CT-RATE (RaTEScore) \\
            \midrule
            w/ Text-Informing & \textbf{73.7} & \textbf{89.8} & \textbf{59.6} \\
            w/o Text-Informing & 72.2 & 86.5 & 57.6 \\
            \bottomrule
        \end{tabular}
        }
    }
\\
\vspace{5pt}
    \subfloat[Effect of Number of Cropped Local Views.]{
            \resizebox{0.42\textwidth}{!}{
            \begin{tabular}{c|ccc}
                \toprule
                \multirow{2}{*}{\textbf{Number of Cropped Views}} & Zero-shot Classification & Segmentation & Report Generation \\
                \cmidrule(lr){2-4}
                 & CT-RATE (AUC) & AMOS (Dice) & CT-RATE (RaTEScore) \\
                \midrule
                1 & 70.4 & 86.4 & 56.5 \\
                2 & 71.8 & 87.9 & 57.1 \\
                3 & \textbf{73.7} & \textbf{89.8} & \textbf{59.6} \\
                4 & 73.0 & 88.3 & 59.1 \\
                \bottomrule
            \end{tabular}
            }
        }
    \caption{Ablation study results for T3D. (a) Comparison of loss functions $\mathcal{L}_{\textrm{GCA}}$ and $\mathcal{L}_{\textrm{TMA}}$. (b) Impact of text-informed alignment in $\mathcal{L}_{\textrm{TMA}}$. (c) Effect of the number of cropped local views used during pre-training. Best results are bolded.}
    \label{tab:ablation_study}
\end{table}

\begin{table}[t]
    \centering
    \resizebox{0.9\linewidth}{!}{
    \begin{tabular}{c|ccc}
                \toprule
                \multirow{2}{*}{\textbf{Number of Transformer Layers}} & Zero-shot Classification & Segmentation & Report Generation \\
                \cmidrule(lr){2-4}
                 & CT-RATE (AUC) & AMOS (Dice) & CT-RATE (RaTEScore) \\
                \midrule
                1 & \textbf{73.7} & \textbf{89.8} & \textbf{59.6} \\
                2 & 73.5 & 89.3 & 59.4 \\
                3 & 73.3 & 89.5 & 59.3 \\
                \bottomrule
            \end{tabular}
        }
        \caption{Performance comparison of models with different transformer layer counts during text-informing for $\mathcal{L}_{\textrm{TMA}}$. Best performance for each task is bolded.}
    \label{tab:layer}
\end{table}

\noindent\textbf{Loss Function Ablation:} We ablate the $\mathcal{L}{\textrm{GCA}}$ and $\mathcal{L}{\textrm{TMA}}$ losses, finding that the best performance is achieved when both losses are used. Removing $\mathcal{L}{\textrm{GCA}}$ reduces performance on the classification task due to the lack of global representation, while removing $\mathcal{L}{\textrm{TMA}}$ significantly harms segmentation and report generation tasks due to the loss of local visual feature learning. These results suggest that using both losses is essential to boost performance across all tasks, as shown in Table \ref{tab:ablation_study}(a).

\noindent\textbf{Text-informed Alignment:} We investigate the impact of incorporating the text-informing strategy in TMA, as shown in Table \ref{tab:ablation_study}(b). Removing text-informing results in a significant drop in performance, particularly for tasks like report generation and segmentation, which rely on learning fine-grained visual features. This decline may be due to the absence of report information, causing the local view embeddings to become ambiguous and harder to associate with their source volume. Without text-informing, the model may confuse regions from different volumes.

\noindent\textbf{Number of Cropped Local Views:} We investigate the impact of the number of local views used during training. Reducing the number of views from 3 to 2 or 1 leads to a decrease in performance, particularly for multi-organ segmentation and report generation tasks. This suggests that a higher number of cropped local views encourages the model to learn more comprehensive spatial features. When increasing the number of views to 4, no further improvement is observed. Based on this, we select 3 local views as the optimal choice for training, as shown in Table \ref{tab:ablation_study}(c).

\noindent\textbf{Model Architecture Hyperparameters:} We ablate the number of transformer layers in the text-informed block \(\mathcal{F}_\psi(\cdot)\), varying the layers from 1 to 3. The results show that performance saturates after a single transformer layer, with minimal improvement observed by adding more layers. This suggests that a single transformer layer is sufficient for text-informed alignment, and further layers do not contribute significantly to the model's performance, as detailed in Table \ref{tab:layer}.

\begin{figure}[h]
    \centering
  \includegraphics[width=0.99\linewidth]{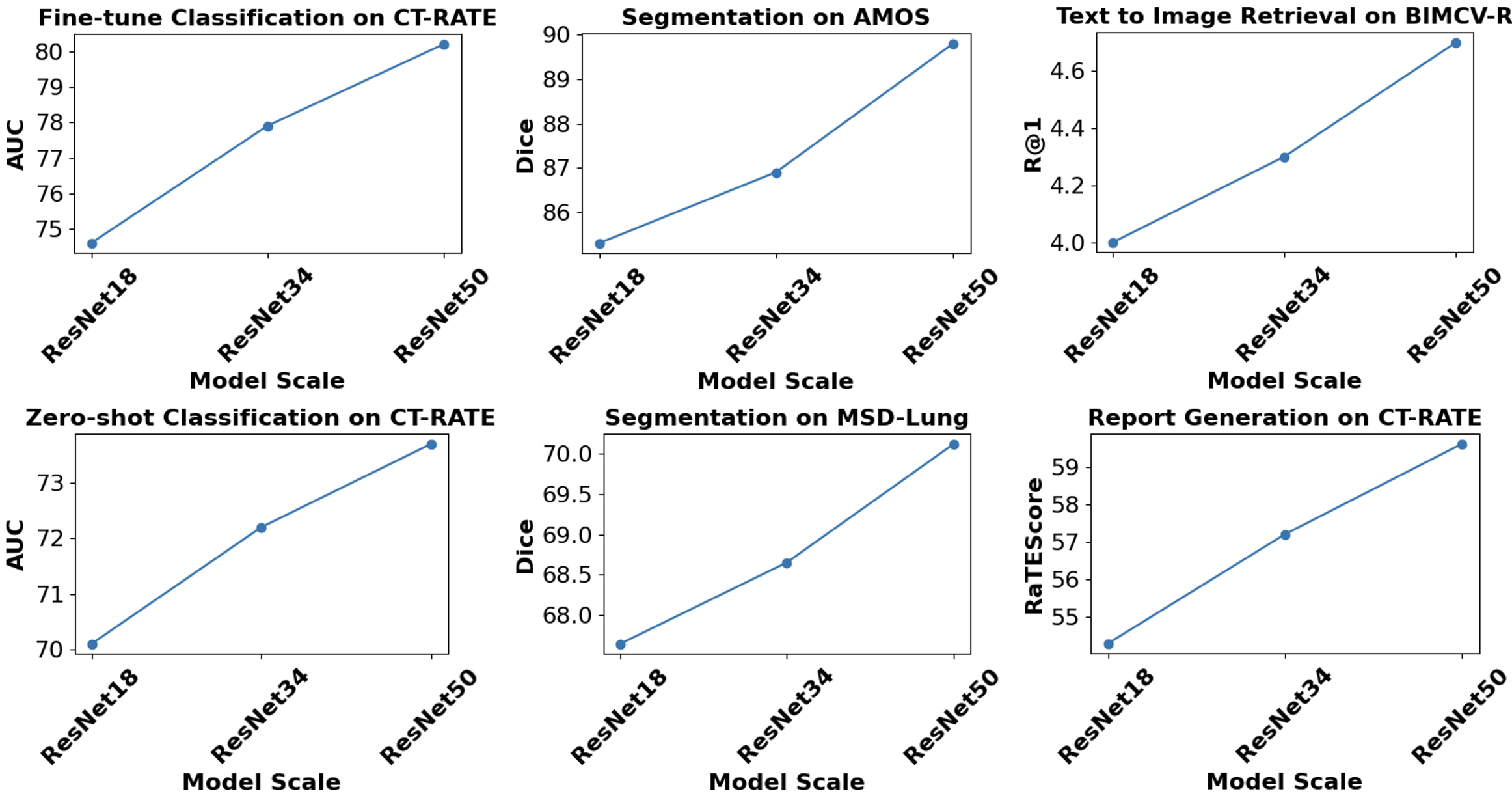}
    \caption{Performance of \textbf{T3D} pre-trained on the proposed \textbf{CT-3DVLP} dataset across six tasks, with varying model scales: ResNet18, ResNet34, and ResNet50. The results show consistent performance improvement as the model scale increases.}
    \label{fig:model scale}
\end{figure}

\noindent\textbf{Model and Data Scalability:} 
We evaluate the impact of both model and data scale on T3D's performance. As shown in Figure \ref{fig:scale}, we analyze the effect of varying pre-training data scales on T3D and CT-CLIP \cite{ct-rate}. Our method consistently surpasses CT-CLIP \cite{ct-rate} from 10k to the full pre-training dataset, demonstrating the effectiveness of T3D across different data scales. Additionally, we evaluate the scalability of our model by testing different ResNet architectures (ResNet18, ResNet34, and ResNet50) as visual encoders. As visualized in Figure \ref{fig:model scale}, T3D shows consistent performance improvements as the model scale increases, highlighting its ability to leverage larger models for better performance across multiple tasks.

\section{Conclusion}
In this work, we present the first and largest publicly available 3D medical VLP dataset, named \textbf{CT-3DVLP}, curated entirely from public resources. We also introduce the \textbf{T3D} framework, which leverages both global alignment and a novel \textbf{text-informed multi-view alignment} strategy to enhance learning and improve performance across various tasks. We demonstrate the effectiveness of T3D on six downstream tasks, including both uni-modal and cross-modal tasks, and show that it outperforms existing methods, such as vSSL and other language supervision approaches that rely on in-house data. Additionally, we highlight the scalability of our method. We believe that T3D, alongside the CT-3DVLP dataset, will make a significant contribution to advancing research in the 3D medical VLP domain.

 \clearpage
\section*{Limitations}
While we propose the T3D framework and the largest publicly available CT-3DVLP dataset, there are several limitations. Even though we have collected nearly all publicly available 3D medical image-report pairs, the dataset still remains limited in size compared to the large-scale datasets used in models like CLIP \cite{radford2021learning}. With only 50k samples, it falls short of the million-level datasets typically used in such models. Additionally, due to the complexities of 3D medical data, it is impractical  to directly leverage powerful 2D visual encoders, limiting the performance of our model. Computational constraints also led us to use ResNet-50 as the visual encoder, rather than more advanced or larger vision models. These limitations point to areas for future work, such as dataset expansion and the integration of more sophisticated 3D vision encoders.

\clearpage
\bibliography{custom}

\clearpage

\appendix

\section{Downstream Task Details}  
\label{sec:downstream detail}

\noindent\textbf{Zero-shot Classification:} We use the CT-RATE dataset \cite{ct-rate} for zero-shot classification, following the protocol in \cite{ct-rate}. T3D is applied without fine-tuning, using the pretrained model for direct classification with the disease name as the category prompt. Evaluation metrics include precision, AUC, accuracy, and F1 score. Image preprocessing is the same as during pretraining.

\noindent\textbf{Fine-tuned Classification:} For fine-tuned classification, we follow the fine-tuning procedure from \cite{ct-rate} on the CT-RATE dataset. The images are preprocessed as in pretraining, and T3D is fine-tuned on the training set and evaluated on the test set. Metrics include accuracy, precision, and recall. We use a batch size of 32, a learning rate of \(1 \times 10^{-3}\), epochs as 50, and cosine learning rate decay. Experiments are run on a single A100-80GB GPU.

\noindent\textbf{Zero-shot Cross-modal Retrieval:} For zero-shot cross-modal retrieval, we use the BIMCV-R dataset \cite{chen2024bimcv} and follow \cite{chen2024bimcv}. Both image and report are embedded into a latent space, and cosine similarity is computed to identify the top-K matches. Retrieval performance is measured using recall@1-10. Image preprocessing is consistent with the pretraining implementation.

\noindent\textbf{Report Generation:} For report generation, we use the official training set from the CT-RATE dataset \cite{ct-rate} and the official test set for evaluation.
Following LLaVA \cite{liu2024visual}, we use Qwen2.5-7b-Instruct as the LLM backbone and the pretrained visual encoder to extract image embeddings. A two-layer MLP serves as the connector, and training is done in two stages: first training the connector, then freezing the ViT and fine-tuning both the connector and the LLM. Generated reports are evaluated using BLEU-1 to 4 and ROUGE-1, 2, L scores. Additional clinical efficacy metrics are adopted from \cite{hamamci2024ct2rep}, with further evaluation using GREEN and RaTEScore \cite{ostmeier2024green, zhao2024ratescore}.

\noindent\textbf{Segmentation Tasks:} For multi-organ segmentation, we use the AMOS \cite{ji2022amos} dataset, following the protocols in \cite{wu2024large}. For lung tumor segmentation, we use the MSD-Lung tumor dataset \cite{msd}. A 3D U-Net architecture is employed, with a pretrained visual encoder and a randomly initialized decoder. Input volumes are normalized to a spacing of 1mm along the three axes, with voxel intensities truncated within the HU range of [-1000, 1000] and normalized to [0,1]. During training, the entire volume is used, with augmentations applied at probabilities of 0.5 for random flipping, 0.3 for rotation, 0.1 for intensity scaling, and 0.1 for shifting. The Dice score is used as the evaluation metric, adhering to the fine-tuning procedure from the official VoCo repository \footnote{https://github.com/Luffy03/VoCo}.

\end{document}